\documentclass[letterpaper]{article} 
\usepackage{aaai2026}  
\usepackage{times}  
\usepackage{helvet}  
\usepackage{courier}  
\usepackage[hyphens]{url}  
\usepackage{graphicx} 
\usepackage{natbib}  
\usepackage{caption} 
\usepackage{algorithm}
\usepackage{algorithmic}
\usepackage{amsmath}
\usepackage{multirow}
\usepackage{amsfonts}
\usepackage{booktabs}
\usepackage{pifont}
\usepackage{bm}
\usepackage{enumitem}
\usepackage{multirow}
\usepackage{newfloat}
\usepackage{listings}
\DeclareCaptionStyle{ruled}{labelfont=normalfont,labelsep=colon,strut=off} 
\floatstyle{ruled}
\newfloat{listing}{tb}{lst}{}
\floatname{listing}{Listing}
\nocopyright 

\title{Stable at Any Speed: Speed-Driven Multi-Object Tracking with Learnable Kalman Filtering}

\author{
    Yan Gong\textsuperscript{\rm 1, \rm 2}, Mengjun Chen\textsuperscript{\rm 3}, Hao Liu\textsuperscript{\rm 2}, Gao Yongsheng\textsuperscript{\rm 1\dag}, Lei Yang\textsuperscript{\rm 4}, Naibang Wang\textsuperscript{\rm 1}, Ziying Song\textsuperscript{\rm 5}, and Haoqun Ma\textsuperscript{\rm 2}
}
\affiliations{
    \textsuperscript{\rm 1}State Key Laboratory of Robotics and System, Harbin Institute of Technology, Harbin, China.\\
    \textsuperscript{\rm 2}Autonomous Driving Department of X Divsion, JD Logistics, Beijing, China.\\
    \textsuperscript{\rm 3}School of Integrated Circuits and Electronics, Beijing Institute of Technology, Beijing, China.\\
    \textsuperscript{\rm 4}School of Mechanical and Aerospace Engineering, Nanyang Technological University, Singapore.\\
    \textsuperscript{\rm 5}Beijing Key Laboratory of Traffic Data Mining and Embodied Intelligence, School of Computer Science and Technology, Beijing Jiaotong University.\\

    Emails: gongyan2020@foxmail.com, fatinn@bit.edu.cn, liuhao163@jd.com, gaoys@hit.edu.cn, lei.yang@ntu.edu.sg, 7788521wnb@gmail.com, songziying@bjtu.edu.cn, mahaoqun1@jd.com

}

\usepackage{bibentry}

\begin{document}

\maketitle

\renewcommand{\thefootnote}{\dag}
\footnotetext{Corresponding Author.}
\renewcommand{\thefootnote}{\arabic{footnote}}


\begin{abstract}
Multi-object tracking (MOT) enables autonomous vehicles to continuously perceive dynamic objects, supplying essential temporal cues for prediction, behavior understanding, and safe planning.
However, conventional tracking-by-detection methods typically rely on static coordinate transformations based on ego-vehicle poses, disregarding ego-vehicle speed–induced variations in observation noise and reference frame changes, which degrades tracking stability and accuracy in dynamic, high-speed scenarios.
In this paper, we investigate the critical role of ego-vehicle speed in MOT and propose a \textbf{S}peed-\textbf{G}uided \textbf{L}earnable \textbf{K}alman \textbf{F}ilter (\textbf{SG-LKF}) that dynamically adapts uncertainty modeling to ego-vehicle speed, significantly improving stability and accuracy in highly dynamic scenarios. Central to SG-LKF is \textbf{M}otion\textbf{S}cale\textbf{Net} (\textbf{MSNet}), a decoupled token-mixing and channel-mixing MLP that adaptively predicts key parameters of SG-LKF. To enhance inter-frame association and trajectory continuity, we introduce a self-supervised trajectory consistency loss jointly optimized with semantic and positional constraints. 
Extensive experiments show that SG-LKF ranks first among all vision-based methods on KITTI 2D MOT with 79.59\% HOTA, delivers strong results on KITTI 3D MOT with 82.03\% HOTA, and outperforms SimpleTrack by 2.2\% AMOTA on nuScenes 3D MOT.
\end{abstract}


\section{Introduction}

Multi-object tracking (MOT) is a fundamental component of autonomous perception, enabling continuous and reliable tracking of dynamic objects to support downstream tasks such as risk prediction and decision making.

Existing 2D and 3D MOT methods are generally categorized as Tracking-by-Detection (TBD) or Joint Detection and Tracking (JDT). While JDT \cite{wang2024multi,tokmakov2021learning, wu2021tracklet,wang2023camo} jointly optimizes detection and tracking for higher accuracy, its complexity limits real-time deployment and modularity. Currently, TBD has become the most widely adopted paradigm, offering greater flexibility and efficiency by decoupling detection and tracking. Within this framework, the classic SORT \cite{bewley2016simple} algorithm exemplifies this approach by combining Kalman filtering (KF) \cite{kalman1960new} with data association, and recent extensions such as DeepSORT \cite{wojke2017simple} and JDE~\cite{wang2020towards} enhance appearance modeling, BoT-SORT \cite{aharon2022bot} fuses motion and appearance cues, and ByteTrack~\cite{zhang2022bytetrack} improves low-score detection association. 
However, these TBD methods employ simplified and fixed-parameter motion models without explicitly accounting for ego-motion or reference frame shifts, thereby compromising tracking stability and causing object loss and identity switches.

\begin{figure*}[!t]
\centering
\includegraphics[width=1.0 \textwidth]{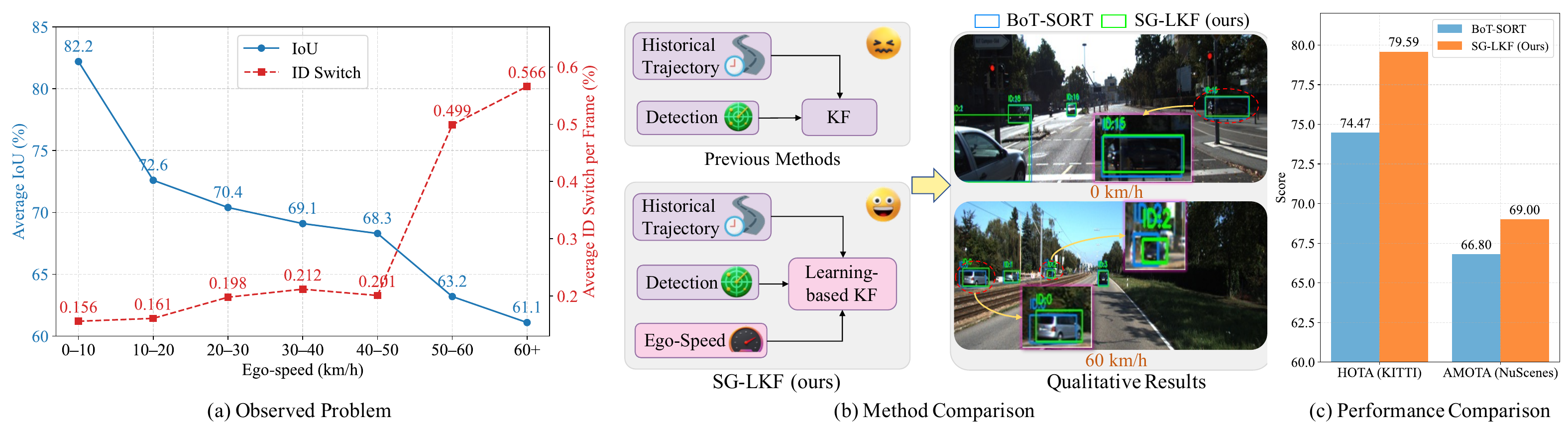} 
\caption{Impact of ego-vehicle speed on multi-object tracking. The proposed SG-LKF achieves precise position prediction in the highlighted red regions of (b) and delivers significant improvements in tracking performance, as shown in (c).}
\label{intro_problem}
\end{figure*}

By analyzing the performance of BoT-SORT on the KITTI 2D MOT dataset under different ego-vehicle speeds (see Fig. \ref{intro_problem} (a)), we observe that as the ego-vehicle speed increases, the Intersection over Union (IoU) between predicted and ground-truth bounding boxes drops significantly, and the ID switch rate rises sharply. This demonstrates that existing tracking methods are highly sensitive to ego-vehicle speed, leading to inaccurate state estimation and association errors for surrounding objects in high-speed driving scenarios. The underlying causes are threefold: (1) \textbf{insufficient reference frame modeling}, where neglecting ego-vehicle motion leads to systematic errors; (2) \textbf{greater inter-frame object displacement at high speeds}, which reduces temporal consistency and increases the difficulty of data association; and (3) \textbf{observation loss from motion blur and occlusion}, forcing the motion model to extrapolate from historical states and thereby exacerbating error accumulation.

To our knowledge, \textbf{we are the first to systematically investigate ego-vehicle speed as a crucial prior for ego-motion modeling}, enabling effective compensation for reference frame shifts caused by ego-vehicle movement and significantly improving state prediction and trajectory continuity. This insight offers a promising direction for addressing the performance bottlenecks of existing multi-object tracking methods under high-speed and rapidly changing driving conditions. 
Furthermore, vehicle speed is a native, real-time signal that requires no additional cost or complexity and is not susceptible to the drift or signal degradation associated with GPS and IMU.

In this paper, we propose a \textbf{S}peed-\textbf{G}uided \textbf{L}earnable \textbf{K}alman \textbf{F}ilter (\textbf{SG-LKF}) that leverages adaptive motion prediction to effectively address dynamic reference frame shifts and consistently deliver stable tracking performance.
Specifically, we design a \textbf{M}otion\textbf{S}caleNet (\textbf{MSNet}) that leverages ego-vehicle speed and object scale as a prior for key parameter generation and hyperparameter adjustment, significantly improving state prediction accuracy. Additionally, we introduce trajectory, position, and semantic consistency loss to enhance object association and trajectory continuity. 
As shown in Fig. \ref{intro_problem} (b), BoT-SORT exhibits more severe bounding box misalignment at 60 km/h, while incorporating ego-vehicle speed in Fig.~\ref{intro_problem} (c) markedly enhances performance.
Experiments show that SG-LKF ranks first among all vision-based methods on the KITTI 2D MOT benchmark, and also achieves leading performance on the KITTI and nuScenes 3D MOT. Moreover, SG-LKF demonstrates strong generalizability and can be seamlessly incorporated into other KF-based tracking frameworks. The main contributions of this paper are summarized as follows:

\begin{itemize}
\item To our knowledge, we are the first to systematically analyze the impact of ego-vehicle speed on multi-object tracking, providing a new perspective and solution.
\item We propose SG-LKF, a speed-guided MOT framework that employs MSNet to adaptively generate key parameters from ego-vehicle speed and object scale, thereby enabling stable tracking performance.
\item We propose a self-supervised trajectory consistency loss, optimized with semantic and positional constraints, to enhance object association and trajectory continuity.
\item Our method ranks first among vision-based methods on the KITTI 2D MOT benchmark, and further demonstrates superior performance on the KITTI and nuScenes 3D MOT datasets.
\end{itemize}

\section{Related Work}



\subsection{Tracking-by-Detection (TBD)}

In recent years, the rapid advancement of object detection techniques \cite{carion2020end,fu2025llmdet,long2025riccardo} has laid a solid foundation for motion-based multi-object tracking methods. 
Mainstream TBD approaches typically employ Kalman filters (KF) to model object motion and focus primarily on the data association problem, which can be broadly categorized into geometry-based and appearance-based methods. Geometry-based~\cite{zhang2022bytetrack,aharon2022bot,maggiolino2023deep} methods use Intersection over Union (IoU) for association, while appearance-based methods~\cite{wojke2017simple,zhang2021fairmot} leverage re-identification (ReID) features to improve robustness under occlusion, interaction, and missing detections. 

\subsection{Motion Modeling in TBD}

In TBD systems, motion modeling is crucial for trajectory continuity. Mainstream approaches typically rely on KF to adopt linear motion assumptions. However, such models struggle to capture complex object dynamics under ego-motion, leading to significant prediction errors and tracking drift. To mitigate this issue, some studies~\cite{du2023strongsort,mahdian2024ego,lin2023asynchronous} incorporate external auxiliary information (e.g., GPS, IMU, or visual odometry) to compensate for ego-motion and improve prediction accuracy. 
However, these methods often require additional hardware or precise calibration, and their effectiveness degrades in challenging environments, such as GPS signal loss in tunnels, IMU drift over time, or visual odometry failures under poor lighting and occlusion.
Therefore, we present the first in-depth analysis of ego-vehicle speed for motion modeling and propose SG-LKF that adaptively generates KF parameters without external sensors, thereby overcoming hardware limitations and ensuring robustness to GPS failure and IMU drift.


\section{Kalman Limitations at High Speeds}
\label{KLAHS}

In conventional KF, the predicted covariance $\mathbf{P}$, the process noise covariance $\mathbf{Q}$, and the observation noise covariance $\mathbf{R}$ are typically preset and remain fixed throughout the tracking process. However, in real-world autonomous driving scenarios, vehicle motion can vary significantly over time. Therefore, standard KF faces significant limitations in high-speed scenarios due to its assumptions of linear dynamics and Gaussian noise, as detailed below.




\noindent \textbf{Nonlinear State Transition.}
The projection from 3D world to 2D image plane follows a nonlinear transformation:
\begin{equation}
    \mathbf{p}_{\text{image}} \propto \frac{\boldsymbol{\Theta}_c(\mathbf{p}_{\text{world}} - \boldsymbol{\tau}_c)}{d_c},
\end{equation}
where $\mathbf{p}_{\text{image}}$ and $\mathbf{p}_{\text{world}}$ are target coordinates on the image plane and in the world, $\boldsymbol{\Theta}_c$ and $\boldsymbol{\tau}_c$ are camera rotation and translation matrix, and $d_c$ is the target depth. When the camera is stationary, this mapping is nearly linear, but at high ego-vehicle speeds, rapid changes in $\boldsymbol{\Theta}_c$ and $\boldsymbol{\tau}_c$ make it highly nonlinear. Nevertheless, the KF uses a fixed linear model:
\begin{equation}
    \hat{\mathbf{x}}_{t|t-1} = \mathbf{F}_t \hat{\mathbf{x}}_{t-1|t-1},
\label{eq:xx}
\end{equation}
where $\hat{\mathbf{x}}_{t|t-1}$ and $\hat{\mathbf{x}}_{t-1|t-1}$ are the predicted and estimated states at $t$ and $t{-}1$, and $\mathbf{F_t}$ denotes the state transition matrix. This mismatch leads to increased prediction error as ego-vehicle speed rises:
\begin{equation}
    \epsilon_{\text{pred}} = \|\hat{\mathbf{x}}_t - \mathbf{x}_t\| \uparrow \quad \text{as ego-vehicle speed } v \uparrow,
\end{equation}
where $\mathbf{x}_t$ is the ground-truth state at time $t$.

\noindent \textbf{Process Noise Covariance Limitation.} 
During high-speed ego-vehicle motion, system uncertainty increases substantially. However, $\mathbf{Q}$ is typically set as a fixed hyperparameter in practice. If $\mathbf{Q}$ is underestimated, the $\mathbf{P}_{t|t-1}$ becomes too small, resulting in a reduced Kalman gain $\mathbf{K}_t$:
\begin{equation}
    \mathbf{K}_t = \mathbf{P}_{t|t-1} \mathbf{H}_t^{\top} \left( \mathbf{H}_t \mathbf{P}_{t|t-1} \mathbf{H}_t^{\top} + \mathbf{R}_t \right)^{-1},
\label{kk}
\end{equation}
where $\mathbf{H_t}$ denotes the observation matrix. As a result, the KF becomes less responsive to new observations, causing errors to accumulate and leading to drift, particularly in high-speed scenarios.

\noindent \textbf{Observation Noise Covariance Limitation.}
In high-speed scenarios, measurement noise increases due to motion blur, rapid viewpoint changes, and occlusions, reducing detection reliability. However, $\mathbf{R}_t$ is typically fixed and cannot reflect this increased uncertainty. As shown in Eq.~\ref{kk}, underestimating $\mathbf{R}_t$ causes the filter to overfit noisy observations, leading to instability, while overestimating it results in underutilized measurements and degraded tracking accuracy.

\section{Methodology}
\subsection{Overview}

\begin{figure*}[!t]
\centering
\includegraphics[width=1.0 \textwidth]{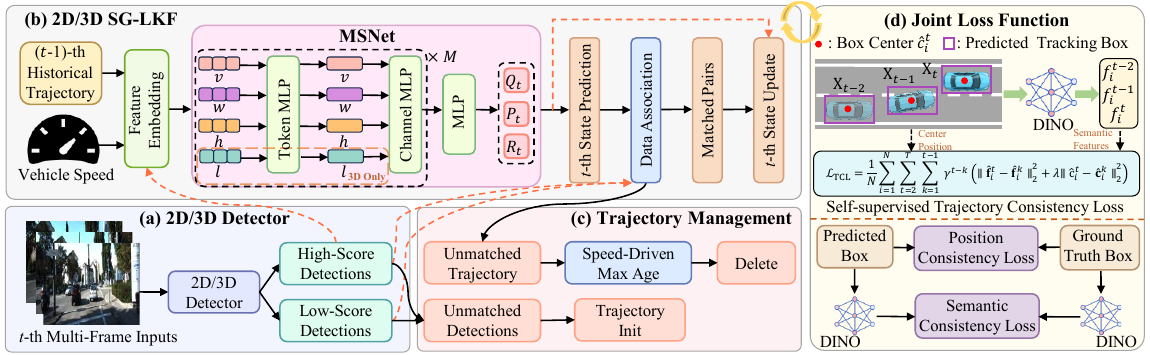} 
\caption{The overall architecture comprises: (a) a 2D/3D detector; (b) a 2D/3D SG-LKF for adaptive motion modeling and state estimation; (c) trajectory management for handling object tracks; and (d) a joint loss function for end-to-end optimization.}
\label{method}
\end{figure*}

To tackle the aforementioned challenges, we reveal the significance of ego-vehicle speed in state estimation and data association, and propose a SG-LKF, as illustrated in Fig.~\ref{method}.
Built upon the standard TBD paradigm, SG-LKF leverages historical trajectories and ego-vehicle speed for state prediction, followed by a two-stage association with high- and low-confidence observations from detectors to enable accurate multi-object tracking.
For matched predictions, SG-LKF adaptively updates states using ego-vehicle speed and observations, and unmatched detections are initialized as new tracks.
Track lifetime is dynamically adjusted based on ego-vehicle speed via linear scaling to prevent unnecessary retention.
Additionally, we introduce a joint loss function to enforce trajectory, spatial, and semantic consistency. The pseudo-code of our SG-LKF is provided in \textit{Appendix A.}


\subsection{Speed-Guided Learnable Kalman Filter (SG-LKF)}

\subsubsection{Learnable Parameterization of $\mathbf{P}$, $\mathbf{Q}$, and $\mathbf{R}$.}

In contrast to existing methods such as KalmanNet \cite{revach2022kalmannet} and HybridTrack \cite{di2025hybridtrack}, which directly incorporate the Kalman gain or the entire state prediction into the learning process, our method preserves the original structure of the KF to the greatest extent, and employs a lightweight learning strategy to adaptively model $\mathbf{P}$, $\mathbf{Q}$, and $\mathbf{R}$. This method retains the theoretical strengths of the KF and introduces a ego-speed-guided adaptation, enabling robust tracking and seamless integration with existing KF-based tracking frameworks.

Following BoT-SORT, we define the state as $\mathbf{x} = [x^\mathbf{x}, y^\mathbf{x}, w^\mathbf{x}, h^\mathbf{x}, \dot{x}^\mathbf{x}, \dot{y}^\mathbf{x}, \dot{w}^\mathbf{x}, \dot{h}^\mathbf{x}]^\top$, and the observation at frame $t$ as $\mathbf{z}_t = [x^\mathbf{z}, y^\mathbf{z}, w^\mathbf{z}, h^\mathbf{z}]^\top$, where ($x$, $y$) denote the 2D coordinates of the object center in the image plane, $\dot{}$ denotes the first-order time derivative, and $w$ and $h$ represent the width and height of the bounding box, respectively. For $\mathbf{P}$, $\mathbf{Q}$ and $\mathbf{R}$, we respectively learn their mappings $f_P$, $f_Q$, and $f_R$ with respect to the current ego-vehicle speed $v$. To enhance adaptability to object scale variations, the $w$ and $h$ are also incorporated as input features.



To capture the relationship between ego-vehicle speed, object scale, and KF parameters, the mapping must be nonlinear, and lightweight for real-time deployment.
Accordingly, we propose MotionScaleNet (MSNet), a decoupled token- and channel-mixing Multilayer Perceptron (MLP) architecture that explicitly models and captures the nonlinear interaction between ego-vehicle speed and object scale. Specifically, given input features $\mathbf{N_0}$=[$v$, $w$, $h$], we first project them into a $\mathbf{C}$-dimensional feature space to form three token sequences. MSNet consists of $L$ ($L=2M$) MLP layers that alternate between token-mixing and channel-mixing operations after LayerNorm (LN) to facilitate cross-semantic information interaction, defined as follows:
\begin{equation}
    \mathbf{N}_L =
    \begin{cases}
        \mathbf{N}_{L-1} + \operatorname{MLP}_{\text{token}}\big(\mathrm{LN}(\mathbf{N}_{L-1})\big), & \text{if $L$ is odd} \\
        \mathbf{N}_{L-1} + \operatorname{MLP}_{\text{channel}}\big(\mathrm{LN}(\mathbf{N}_{L-1})\big), & \text{if $L$ is even}
    \end{cases}
\end{equation}
Finally, $\mathbf{N}_L$ is flattened and passed through a regression MLP with softplus activation to predict positive semi-definite KF parameters.

\subsubsection{Adaptive Kalman Filtering Process.} At time step $t$, KF operates in two stages: \emph{prediction} and \emph{update}. In the \textit{prediction} stage, process noise covariance $\mathbf{Q}_t$ is first adaptively generated by MSNet using $v_{t-1}$, $w_{t-1}^\mathbf{x}$, and $h_{t-1}^\mathbf{x}$ to capture the inherent uncertainty in the motion model. The state $\hat{\mathbf{x}}_{t|t-1}$ is estimated from $\hat{\mathbf{x}}_{t-1|t-1}$ in Eq. \ref{eq:xx}, and covariance $\mathbf{P}_{t|t-1}$ is estimated from $\mathbf{Q}_t$ and $\mathbf{P}_{t-1|t-1}$, as defined by:

\begin{align}
    &\textit{Prediction:} \left\{
    \begin{aligned}
    &\mathbf{Q}_{t} = \text{MSNet}(v_{t-1}, w^\mathbf{x}_{t-1}, h^\mathbf{x}_{t-1}) \\
        &\mathbf{P}_{t|t-1} = \mathbf{F}_t \mathbf{P}_{t-1|t-1} \mathbf{F}_t^\top + \mathbf{Q}_t,
    \end{aligned}
    \right.
\end{align}
In the \textit{update} stage, the MSNet first generate $\mathbf{R}_t$ and $\mathbf{P}_{t|t}$ using $v_t$, $w_t^\mathbf{z}$, and $h_t^\mathbf{z}$. Subsequently, $\mathbf{K}_t$ is computed from $\mathbf{R}_t$ and the $\mathbf{P}_{t|t-1}$ in Eq. \ref{kk}, and the state $\hat{\mathbf{x}}_{t|t}$ is updated using $\mathbf{K}_t$, the observation $\mathbf{z}_t$, and the previous state $\hat{\mathbf{x}}_{t|t-1}$, as defined by:

\begin{align}
    &\textit{Update:} \left\{
    \begin{aligned}
        &\mathbf{R}_{t} = \text{MSNet}(v_t, w^\mathbf{z}_t, h^\mathbf{z}_t) \\
        &\mathbf{P}_{t|t} = \text{MSNet}(v_t, w^\mathbf{z}_t, h^\mathbf{z}_t) \\
        &\hat{\mathbf{x}}_{t|t} = \hat{\mathbf{x}}_{t|t-1} + \mathbf{K}_t(\mathbf{z}_t - \mathbf{H} \hat{\mathbf{x}}_{t|t-1}), \\
    \end{aligned}
    \right.
\end{align}

\subsubsection{Extension to 3D Multi-Object Tracking.} To extend the 2D SG-LKF framework to 3D MOT, we replace the 2D detector with a 3D detector for spatial pose-aware detections and redefine the KF state as $\mathbf{x} = [x^\mathbf{x}, y^\mathbf{x}, z^\mathbf{x}, w^\mathbf{x}, h^\mathbf{x}, l^\mathbf{x}, \dot{x}^\mathbf{x}, \dot{y}^\mathbf{x}, \dot{z}^\mathbf{x}, \dot{w}^\mathbf{x}, \dot{h}^\mathbf{x}, \dot{l}^\mathbf{x}]^\top$, where $z^\mathbf{x}$ and $l^\mathbf{x}$ denote the vertical position of the object center and the length. Similarly, MSNet incorporates object length $l$ into the KF parameter prediction to enhance scale-aware motion modeling.
For data association, we adopt the geometry-aware similarity metric from SimpleTrack~\cite{pang2022simpletrack} and leverage voxel-based dynamic point cloud aggregation~\cite{wu2023virtual} to enhance structural representation and tracking stability.

\subsection{Loss Function Design}
\noindent \textbf{Trajectory Consistency Loss (TCL).} 
To mitigate trajectory inconsistency from detection noise and motion blur in high-speed scenarios, we propose a TCL to enforce temporal smoothness and appearance coherence.
We define a trajectory for object $i$ over $T$ frames as a sequence of predicted states: $\mathcal{T}_i = \{(\hat{\mathbf{c}}_i^t, \hat{\mathbf{f}}_i^t)\}_{t=1}^{T}$, where $\hat{\mathbf{c}}_i^t \in \mathbb{R}^2$ denotes the predicted 2D center position, and $\hat{\mathbf{f}}_i^t \in \mathbb{R}^d$ represents the semantic embedding of the predicted bounding box at frame $t$. We assume that objects along the same trajectory should exhibit consistent semantics and smooth positional transitions across consecutive frames, defined as follows: $ \left\| \hat{\mathbf{f}}_i^t - \hat{\mathbf{f}}_i^{t-1} \right\|_2^2 + \lambda \cdot \left\| \hat{\mathbf{c}}_i^t - \hat{\mathbf{c}}_i^{t-1} \right\|_2^2$, where $\|\cdot\|_2$ and $\lambda$ denote the L2 norm and the weighting factor balancing positional and semantic consistency, respectively. However, computing trajectory consistency using only adjacent frames may introduce errors due to potential semantic shifts or positional deviations. To mitigate this, we introduce a temporal accumulation mechanism that emphasizes long-term consistency and reduces the impact of local noise, defined as follows: 
\begin{equation}
    \mathcal{L}_{\mathrm{TCL}}=\frac{1}{N} \sum_{i=1}^N \sum_{t=2}^T \sum_{k=1}^{t-1} \gamma^{t-k}\left(\left\|\hat{\mathbf{f}}_i^t-\hat{\mathbf{f}}_i^k\right\|_2^2+\lambda\left\|\hat{\mathbf{c}}_i^t-\hat{\mathbf{c}}_i^k\right\|_2^2\right),
\end{equation}
where $\hat{\mathbf{f}}_i^k$ and $\hat{\mathbf{c}}_i^k$ are the Mamba-inspired \cite{gu2023mamba} temporal aggregates over the past $k$ steps. $N$ denotes the number of tracked trajectories. $\gamma^{t-k}$ applies exponential decay to prioritize nearby-frame consistency while retaining sensitivity to long-range temporal variations. 


\noindent \textbf{Semantic Consistency Loss.} 
To enforce semantic alignment between predictions and ground truths, we extract semantic embeddings 
$\hat{\mathbf{f}}_i$ and $\mathbf{f}_i$ from the predicted and corresponding ground-truth bounding boxes, and compute their cosine similarity, defined as $\mathcal{L}_{\mathrm{SCL}}$.

\noindent \textbf{Position Consistency Loss.} To ensure precise spatial alignment between predictions and ground truth, we employ the Complete IoU (CIoU) loss \cite{zheng2021enhancing}, which accounts for overlap area and center distance, defined as $\mathcal{L}_{\mathrm{PCL}}$. 

\noindent \textbf{Overall Loss.} The total loss combines trajectory, semantic, and position consistency terms as a weighted combination to jointly enforce appearance alignment, spatial accuracy, and temporal smoothness, defined as: 
\begin{equation}
    \mathcal{L} = \mathcal{L}_{\mathrm{TCL}} + \alpha \cdot \mathcal{L}_{\mathrm{SCL}} + \beta \cdot \mathcal{L}_{\mathrm{PCL}},
\end{equation}
where $\alpha$ and $\beta$ denote weighting factors.

\section{Experiment}

\subsection{Datasets and Evaluation Metrics}
\textbf{KITTI} \cite{geiger2012we} is a widely used benchmark for 2D and 3D multi-object tracking (MOT), featuring 21 training and 29 test sequences across urban, residential, and highway scenes. 

\noindent
\textbf{nuScenes} \cite{caesar2020nuscenes} is a large-scale autonomous driving dataset comprising 1,000 urban and suburban scenes, featuring 6 cameras, 5 radars, a 32-line LiDAR, and 1.4 M annotated 3D bounding boxes across 23 categories, along with HD maps and vehicle sensor data.

\noindent
\textbf{Evaluation Metrics.}
For KITTI, we adopt Higher Order Tracking Accuracy (HOTA) as the primary metric, and additionally report Detection Accuracy (DetA), Association Accuracy (AssA), Detection Recall (DetRe), Detection Precision (DetPr), Association Recall (AssRe), Association Precision (AssPr), and Localization Accuracy (LocA).
For nuScenes 3D tracking, we use Average Multi-Object Tracking Accuracy (AMOTA) as the main metric, along with Average Multi-Object Tracking Precision (AMOTP), Multi-Object Tracking Accuracy Recall (MOTAR), Multi-Object Tracking Accuracy (MOTA), Multi-Object Tracking Precision (MOTP), and Recall (REC). All metrics are reported as percentages throughout this paper.

\begin{table}[!t]
\small
\centering
\setlength{\tabcolsep}{3pt} 
\begin{tabular}{ccccccc}
\toprule
\textbf{SG-LKF} & $\bm{\mathcal{L}_{\mathrm{PCL}}}$  & $\bm{\mathcal{L}_{\mathrm{SCL}}}$ &  $\bm{\mathcal{L}_{\mathrm{TCL}}}$ & \textbf{HOTA} $\uparrow$            & \textbf{DetA} $\uparrow$           & \textbf{AssA} $\uparrow$            \\
\midrule
             &            &               &                 & 74.47          & 75.48          & 74.30           \\
           \ding{51}  &     \ding{51}       &               &                 & 76.84          & 77.08           & 77.24          \\
           \ding{51}  &       \ding{51}     &   \ding{51}            &                 & 79.49          & \textbf{77.81} & 81.81          \\
          \ding{51}   &      \ding{51}      &               &    \ding{51}             & 79.22          & 77.77           & 81.29          \\
         \ding{51}    &      \ding{51}     &  \ding{51}             &            \ding{51}     & \textbf{79.59} & 77.27          & \textbf{82.53} \\
\bottomrule
\end{tabular}
\caption{Comparison of different components in our method on KITTI 2D MOT. The best is marked in bold. $\uparrow$ indicates higher is better, while $\downarrow$ indicates lower is better.}
\label{tab:component_ablation}
\end{table}



\begin{figure}[!t]
\centering
\includegraphics[width=1.0 \linewidth]{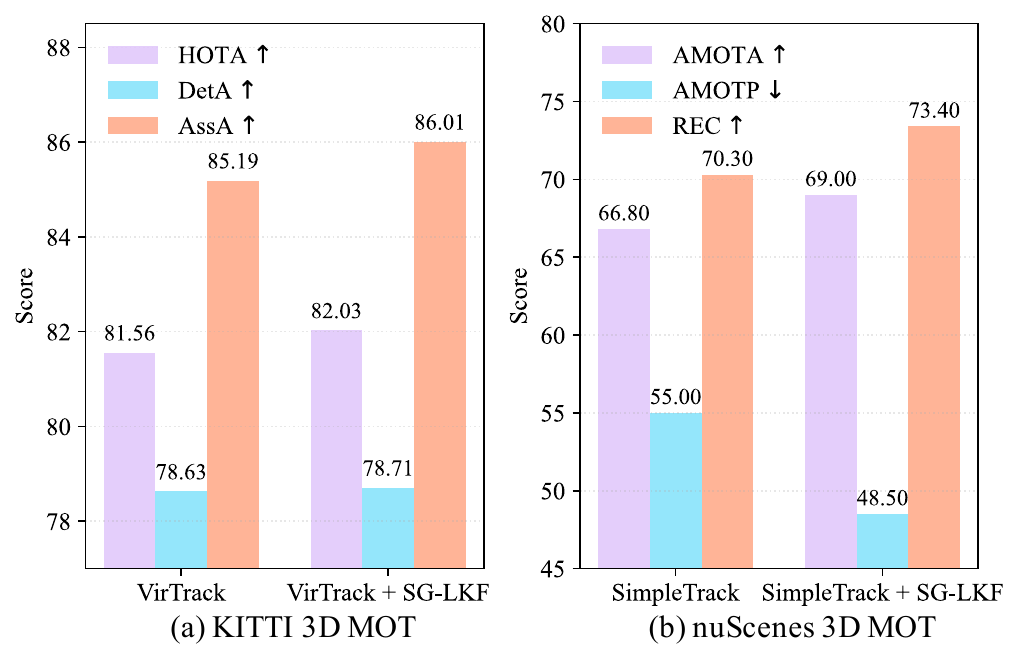} 
\caption{Comparison of performance improvements achieved by integrating SG-LKF into VirConvTrack on KITTI 3D MOT and SimpleTrack on nuScenes 3D MOT.}
\label{fig:zhuzhuangtu}
\end{figure}

\subsection{Implementation details}
Our model is implemented in PyTorch and trained end-to-end on a single NVIDIA GeForce RTX 3090 GPU (24GB). The optimization is performed using AdamW with an initial learning rate of 5e-3 and weight decay of 1e-2. A learning rate schedule with linear warm-up over the first 5 epochs is followed by cosine annealing decay over 100 total epochs.
The $L$ of MSNet is set to 6.
Notably, the loss weights are treated as learnable parameters and are jointly optimized with the network.
We use DINOv2~\cite{oquab2023dinov2} to extract semantic features of bounding boxes only during loss computation, avoiding ReID features at inference.
Moreover, MSNet is extremely lightweight, with only 78.2 K parameters and 79.5 K FLOPs.

\subsection{Ablation Study}
\noindent \textbf{Component Ablation.} Table~\ref{tab:component_ablation} reports the contribution of four core components—SG-LKF, $\mathcal{L}_{\mathrm{PCL}}$, $\mathcal{L}_{\mathrm{SCL}}$, and $\mathcal{L}_{\mathrm{TCL}}$—to the overall performance. Compared to the baseline (BoT-SORT) in the first row, introducing SG-LKF leads to notable gains of 2.37\%, 1.60\%, and 2.94\% in HOTA, DetA, and AssA, respectively, indicating that SG-LKF effectively compensates for ego-motion-induced reference shifts and improves both prediction accuracy and association robustness.
Further, $\mathcal{L}_{\mathrm{SCL}}$ and $\mathcal{L}_{\mathrm{TCL}}$ individually improve HOTA by 2.65\% and 2.38\%, while their combination achieves the best overall performance, highlighting the complementary roles of semantic and temporal consistency in improving identity preservation and trajectory continuity. Further ablation studies on $\mathbf{P}$, $\mathbf{Q}$, and $\mathbf{R}$ are provided in \textit{Appendix B.}

\begin{figure}[!t]
\centering
\includegraphics[width=1.0 \linewidth]{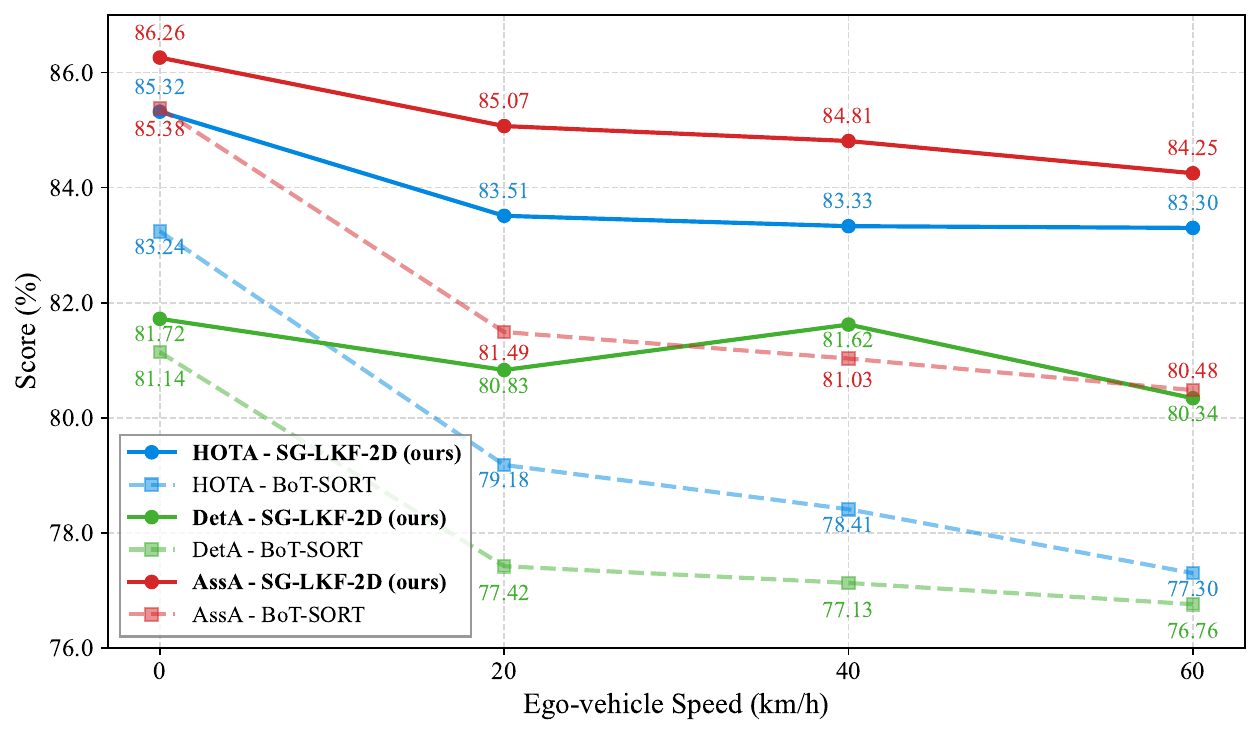} 
\caption{Comparison of BoT-SORT and SG-LKF-2D (ours) at different ego-vehicle speeds on KITTI 2D MOT.}
\label{fig:velocity_curve}
\end{figure}


\begin{figure}[!t]
\centering
\includegraphics[width=1.0 \linewidth]{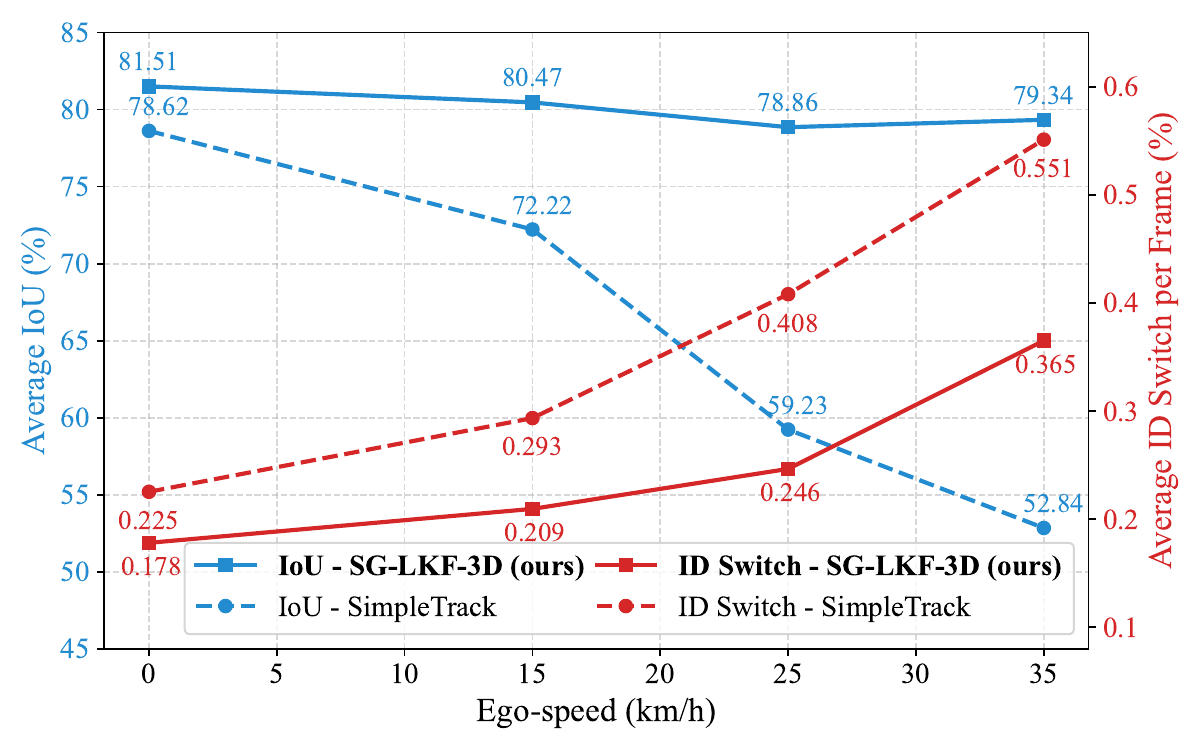} 
\caption{Comparison of SimpleTrack and SG-LKF-3D (ours) at different ego-vehicle speeds on nuScenes 3D MOT.}
\label{fig:iou_idswitch_nus}
\end{figure}

\begin{table*}[!ht]
\centering
\resizebox{\linewidth}{!}{
\begin{tabular}{lccccccccc}
\toprule
\textbf{Method}                & \textbf{Speed Information} & \textbf{HOTA} $\uparrow$   & \textbf{DetA} $\uparrow$   & \textbf{AssA} $\uparrow$   & \textbf{DetRe} $\uparrow$  & \textbf{DetPr} $\uparrow$  & \textbf{AssRe} $\uparrow$  & \textbf{AssPr} $\uparrow$  & \textbf{LocA} $\uparrow$   \\
\midrule
BoT-SORT              &             & 74.47  & 75.48  & 74.30  & 79.91  & 84.15  & 78.44  & 86.52  & 86.14  \\
\midrule
\multirow{5}{*}{\textbf{SG-LKF-2D (ours)}} &  $v$       & \textbf{79.59}  & 77.27  & \textbf{82.53}  & 80.55  & \textbf{86.58}  & \textbf{85.23}  & 89.79  & \textbf{87.09}  \\
                      & $v \left(1 \pm \text{5\%} \cdot \mathcal{N}(0, 1) \right)$     & 79.56  & 77.26  & 82.49  & 80.53  & 86.57  & 85.19  & 89.77  & 87.08  \\
                      & $v \left(1 \pm \text{10\%} \cdot \mathcal{N}(0, 1) \right)$      & 79.53  & \textbf{77.68}  & 82.13  & \textbf{81.43}  & 85.89  & 85.05  & \textbf{89.81}  & 86.95  \\
                      & $v \left(1 \pm \text{20\%} \cdot \mathcal{N}(0, 1) \right)$      & 79.27  & 77.17  & 82.00  & 80.54  & 86.45  & 84.73  & 89.76  & \textbf{87.09}  \\
                      & $v  \cdot \mathcal{N}(0, 1)$       & 78.84  & 77.50  & 80.77  & 81.05  & 86.26  & 83.45  & 89.87  & 87.04  \\
\bottomrule
\end{tabular}
}
\caption{Robustness evaluation of SG-LKF-2D under varying levels of Gaussian White Noise $\mathcal{N}(0, 1)$ on KITTI 2D MOT.}
\label{tab:robustness}
\end{table*}

\noindent \textbf{2D-to-3D MOT with Ego-Vehicle Speed Modeling}. Fig~\ref{fig:zhuzhuangtu} validates the effectiveness of SG-LKF in 3D multi-object tracking. On KITTI, integrating SG-LKF into VirConvTrack~\cite{wu2023virtual} improves HOTA by 0.47\%, DetA by 0.08\%, and AssA by 0.82\%. On nuScenes, incorporating SG-LKF into SimpleTrack~\cite{pang2022simpletrack} boosts AMOTA by 2.2\%, lowers AMOTP by 6.5\%, and increases REC by 3.1\%. 
These gains demonstrate that SG-LKF enhances motion modeling and trajectory continuity, while generalizing well across diverse baselines.

\noindent \textbf{Quantitative Sensitivity to Ego-Vehicle Speed}. To assess the impact of ego-vehicle speed on MOT performance, we evaluate both BoT-SORT and SG-LKF-2D across varying speeds (0, 20, 40, and 60 km/h) in terms of HOTA, DetA, and AssA, as illustrated in Fig. \ref{fig:velocity_curve}. Two key observations emerge: (1) SG-LKF-2D consistently outperforms the baseline across all speeds and metrics, with particularly notable gains at higher speed; (2) while both methods exhibit performance degradation as speed increases, SG-LKF-2D maintains significantly greater stability, whereas the baseline suffers a sharp decline.
Fig.~\ref{fig:iou_idswitch_nus} compares SimpleTrack and SG-LKF-3D on the nuScenes 3D MOT benchmark under specific speed cases (0, 15, 25, and 35 km/h), demonstrating that SG-LKF-3D consistently achieves higher IoU and fewer ID switches.
These results clearly demonstrate the superior adaptability of SG-LKF to dynamic motion and highlight the critical role of ego-vehicle speed modeling in mitigating reference frame shifts and preserving identity consistency. 

\noindent \textbf{Robustness to Speed Noise}.
In real-world autonomous driving, speed measurements are inevitably affected by environmental disturbances and system errors. To assess SG-LKF’s robustness to speed noise, we injected Gaussian white noise of varying intensities into the input. As shown in Table~\ref{tab:robustness}, SG-LKF-2D maintains stable performance across increasing noise levels and consistently outperforms the baseline. These results confirm its robustness to speed perturbations, ensuring reliable tracking under challenging conditions.

\subsection{Benchmarks Evaluation}

\begin{table}[!t]
\centering
\setlength{\tabcolsep}{1pt} 

\resizebox{\linewidth}{!}{
\begin{tabular}{lccccccc}
\toprule
\textbf{Track} & \textbf{HOTA $\uparrow$}   & \textbf{DetA $\uparrow$}   & \textbf{AssA $\uparrow$} & \textbf{DetRe $\uparrow$}  & \textbf{DetPr $\uparrow$} & \textbf{AssRe $\uparrow$} & \textbf{AssPr $\uparrow$} \\
\midrule

CenterTrack   & 73.02           & 75.62           & 71.20         & 80.10           & 84.56          & 73.84          & 89.00                   \\
LGM            & 73.14           & 74.61           & 72.31         & 80.53           & 82.16          & 76.38          & 84.74                     \\
FNC2           & 73.19           & 73.27           & 73.77         & 80.98           & 81.67          & 77.05          & \textbf{89.84}                 \\
DEFT           & 74.23           & 75.33           & 73.79         & 79.96           & 83.97          & 78.30          & 85.19              \\
APPTracker+    & 75.19           & 75.55           & 75.36         & 78.77           & 86.04          & 78.34          & 88.24             \\
OC-SORT        & 76.54           & 77.25           & 76.39         & 80.64           & 86.36          & 80.33          & 87.17                 \\
S3MOT          & 76.86             & 76.95             & 77.41           & 83.79             &       $-$           & 81.01            & $-$                          \\
C-TWiX         & 77.58           & 76.97           & 78.84         & 80.25           & 86.43          & 81.90          & 88.35                   \\
PermaTrack     & 78.03           & 78.29           & 78.41         & 81.71           & \underline{86.54}          & 81.14          & 89.49                 \\
JHIT           & 79.21           & 76.76           & \underline{82.29}         & 81.63           & 84.62          & \textbf{85.94}          & 88.19                   \\
RBMOT          & 79.23             & 76.07             & 83.13           & $-$                  &       $-$           &      $-$            &    $-$                              \\
RAM            &79.53            &\textbf{78.79}  & 80.94         & \textbf{82.54} & 86.33          & 84.21          & 88.77         \\
 \textbf{SG-LKF-2D (ours)}  &\textbf{79.59} & \underline{77.27} & \textbf{82.53}         & \underline{80.55} & \textbf{86.58}          & \underline{85.23}          & \underline{89.79}         \\
\bottomrule
\end{tabular}
}
\caption{Comparison of SG-LKF-2D with existing methods on the KITTI 2D MOT test set. \underline{Underline} marks the second-best. ``$-$'' denotes results not reported in the original paper.}
\label{tab:kitti_2D}
\end{table}

\subsubsection{KITTI 2D Results.}
We compare SG-LKF-2D with a comprehensive set of state-of-the-art multi-object tracking methods, including CenterTrack \cite{zhou2020tracking}, LGM \cite{wang2021track}, FNC2 \cite{jiang2023novel}, DEFT \cite{chaabane2021deft}, APPTracker+ \cite{zhou2024apptracker+}, OC-SORT \cite{cao2023observation}, S3MOT \cite{yan2025s3mot}, C-TWiX \cite{miah2025learning}, PermaTrack \cite{tokmakov2021learning}, JHIT \cite{claasen2024interacting}, RBMOT \cite{liu2025adaptive}, and RAM \cite{tokmakov2022object}. As shown in Table~\ref{tab:kitti_2D}, SG-LKF-2D achieves the highest HOTA of 79.59\%, outperforming all other methods. Notably, SG-LKF-2D achieves the best or second-best results across all metrics. 
Since we directly adopt the detection results from BoT-SORT, DetA is slightly lower than that of RAM.
These results demonstrate that the effective modeling of speed information in SG-LKF markedly improves both data association and tracking stability.

\begin{table}[!t]
\centering
\setlength{\tabcolsep}{1pt} 
\resizebox{\linewidth}{!}{
\begin{tabular}{lccccccc}
\toprule
\textbf{Methods} & \textbf{HOTA $\uparrow$}   & \textbf{DetA $\uparrow$}   & \textbf{AssA $\uparrow$}   & \textbf{DetRe $\uparrow$}  & \textbf{DetPr $\uparrow$}  & \textbf{AssRe $\uparrow$}  & \textbf{AssPr $\uparrow$}     \\
\midrule
Stereo3DMOT      & 77.32           & 73.43           & 81.86           & 77.34           & 85.38           & 84.66           & 89.61                  \\
PC3T             & 77.80           & 74.57           & 81.59           & 79.19           & 84.07           & 84.77           & 88.75                   \\
MSA-MOT          & 78.52           & 75.19           & 82.56           & 82.42           & 82.21           & 85.21           & 90.16                      \\
UG3DMOT          & 78.60           & 76.01           & 82.28           & 80.77           & 85.44           & 85.36           & 91.37                      \\
YONTD-MOTv2      & 79.52           &     75.83               & 84.01           & 82.31           & 83.69           & 87.16           & 90.70                   \\
ReThink\_MOT      & 80.39           & 77.88           & 83.64           & \underline{84.23}           & 83.57           & 87.63           & 88.90                    \\
LEGO             & 80.75           & \textbf{78.91}           & 83.27           & \textbf{84.64}           & 84.94           & 86.87           & 90.19                     \\
PC-TCNN          & 80.90           & 78.46           & 84.13           & 84.22           & 84.58           & 87.46           & 90.47                   \\
PMM-MOT          & 80.27           & $-$           & 83.12           & $-$            & $-$            & $-$            & $-$                    \\
PMTrack          & 81.36             & 78.90             & 84.49           & 88.02                  &       $-$           &      $-$            &    $-$                              \\
MCTrack\_online  & 81.07           & 78.08           & \underline{84.82}           & 83.63           & 85.09           & 87.61           & \underline{91.43}                     \\
RobMOT\_CasA     & 81.22           & 78.34           & 84.80           & 81.95           & \textbf{87.21}           & 87.40           & \textbf{91.44}                   \\
VirConvTrack     & \underline{81.56}           & 78.63           & \underline{85.19}           & 82.39           & 87.13           & \underline{88.70}           & 90.49                     \\
 \textbf{SG-LKF-3D (ours)}    & \textbf{82.03} & \underline{78.71} & \textbf{86.01} & 82.47 & \underline{87.16} & \textbf{88.79} & 90.55 \\
\bottomrule
\end{tabular}
}
\caption{Comparison of SG-LKF-3D with existing methods on the KITTI 3D MOT test set.}
\label{tab:kitti_3D}
\end{table}

\subsubsection{KITTI 3D Results.}
On the KITTI 3D MOT test set, we compare SG-LKF-3D with a range of existing methods, including Stereo3DMOT \cite{mao2023stereo3dmot}, PC3T \cite{wu20213d}, MSA-MOT \cite{zhu2022msa}, UG3DMOT \cite{he20243d}, YONTD-MOTv2 \cite{wang2024multi}, ReTink\_MOT \cite{wang2023towards}, LEGO \cite{zhang2023lego}, PC-TCNN \cite{wu2021tracklet}, PMM-MOT~\cite{gohari2025adaptive}, PMTrack~\cite{tan20253d}, MCTrack\_online \cite{wang2024mctrack}, RobMOT\_CasA \cite{nagy2024robmot}, and VirConvTrack \cite{wu2023virtual}. As shown in Table~\ref{tab:kitti_3D}, SG-LKF-3D achieves a HOTA of 82.03\%, outperforming all existing methods and demonstrating the best overall performance. SG-LKF-3D attains the highest scores in key association metrics, with AssA (86.01\%) and AssRe (88.79\%).
Overall, these results demonstrate that SG-LKF-3D delivers superior 3D multi-object tracking accuracy, further validating the effectiveness of our speed modeling.


\begin{table}[!t]
\centering
\setlength{\tabcolsep}{1pt} 
\resizebox{\linewidth}{!}{
\begin{tabular}{lcccccc}
\toprule
\textbf{Methods} & \textbf{AMOTA $\uparrow$} & \textbf{AMOTP $\downarrow$} & \textbf{MOTAR $\uparrow$} & \textbf{MOTA $\uparrow$} & \textbf{MOTP $\uparrow$} & \textbf{REC $\uparrow$} \\
\midrule
UG3DMOT         & 66.80   & 53.80   & 79.20   & 54.90   & \underline{31.00}   & 70.00 \\
ImmortalTracker & 67.70   & 59.90      & 80.00               & \underline{57.20}   & 28.50            & 71.40 \\
3DMOTFormer++   & 68.20   & 49.60               & \textbf{81.00}      & 55.60   & 29.70              & 69.30 \\
MLPMOT          & 68.30   & \underline{49.00}               & 77.50               & 55.40   & \textbf{31.10}     & 72.80 \\
Wang et al.          & 48.20   &       $-$         & $-$               & 40.70   &   $-$   & $-$ \\
TransFusion-L   & \underline{68.60}   & 52.90               & 78.40               & 57.10   & \underline{31.00}   & \underline{73.10} \\
SimpleTrack     & 66.80   & 55.00               & \underline{80.80}   & 56.60   & 29.40              & 70.30 \\
 \textbf{SG-LKF-3D (ours)}   & \textbf{69.00}   & \textbf{48.50}      & 79.50               & \textbf{57.60}   & 29.40              & \textbf{73.40} \\
\bottomrule
\end{tabular}
}
\caption{Comparison of SG-LKF-3D with existing methods on the nuScenes 3D MOT test set.}
\label{tab:nus_3D}
\end{table}

\subsubsection{nuScenes 3D Results.}
On the nuScenes 3D MOT test set, we compare SG-LKF-3D with several advanced methods, including UG3DMOT, ImmortalTracker \cite{wang2021immortal}, 3DMOTFormer++ \cite{ding20233dmotformer}, MLPMOT \cite{benbarka2021score}, Wang et al.~\cite{wang2025uncertain}, TransFusion-L \cite{bai2022transfusion}, and SimpleTrack. As shown in Table \ref{tab:nus_3D}, SG-LKF-3D outperforms all competing methods on two key metrics, achieves an AMOTA of 69.00\% and an AMOTP of 48.50\%. SG-LKF-3D also excels in MOTAR and REC, achieving 79.50\% and 73.40\%, respectively, both ranking among the top results. These findings clearly validate the accuracy of SG-LKF-3D in 3D MOT, further highlighting the effectiveness of speed modeling.

\begin{figure}[!t]
\centering
\includegraphics[width=1.0 \linewidth]{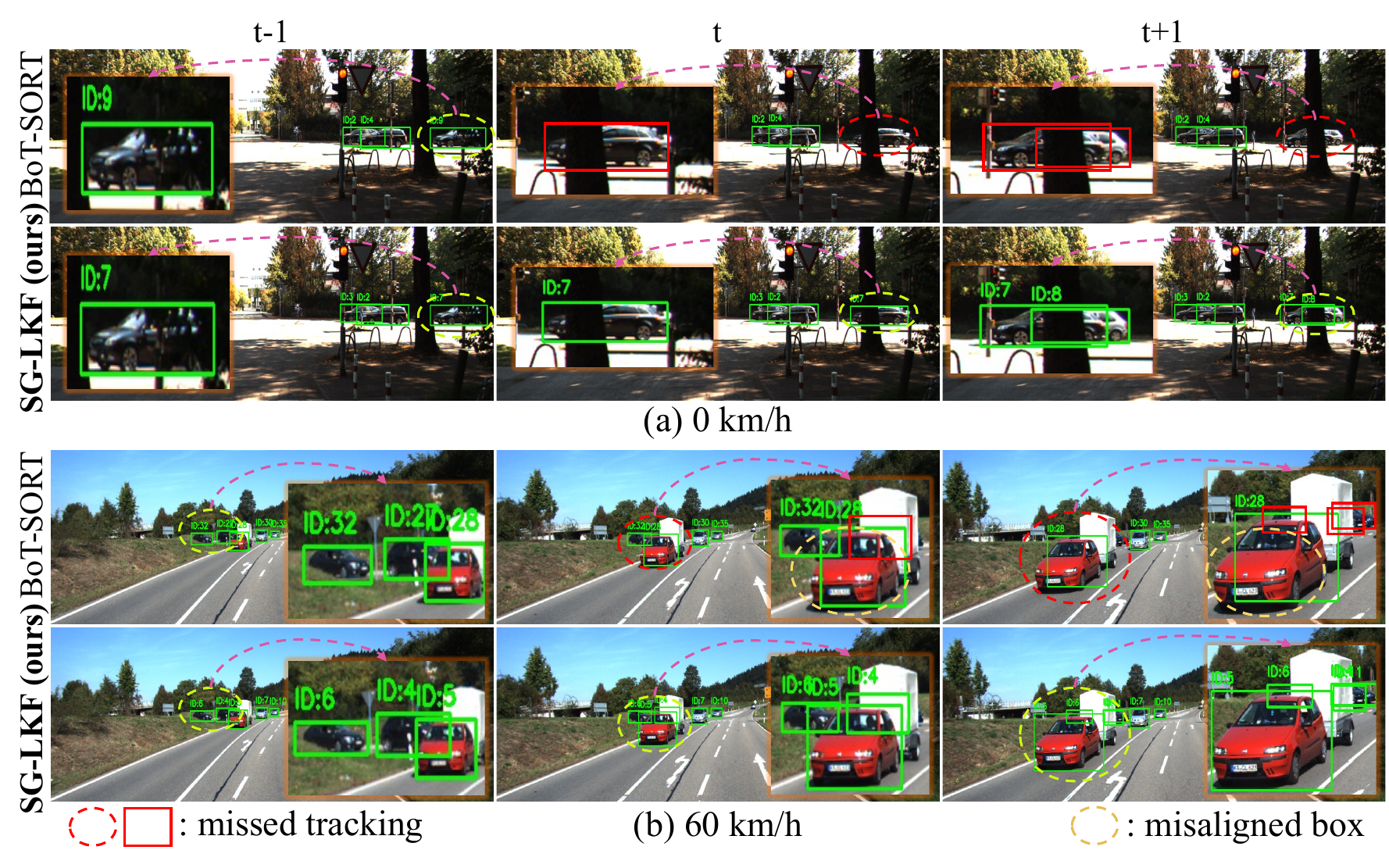} 
\caption{Qualitative comparison of BoT-SORT and SG-LKF-2D (ours) at 0 km/h and 60 km/h on the KITTI 2D MOT test set. At 0 km/h, tree occlusion causes missed tracking in the baseline; at 60 km/h, it produces misaligned boxes and misses occluded vehicles.}
\label{fig:vis_result_kitti}
\end{figure}

\begin{figure}[!t]
\centering
\includegraphics[width=1.0 \linewidth]{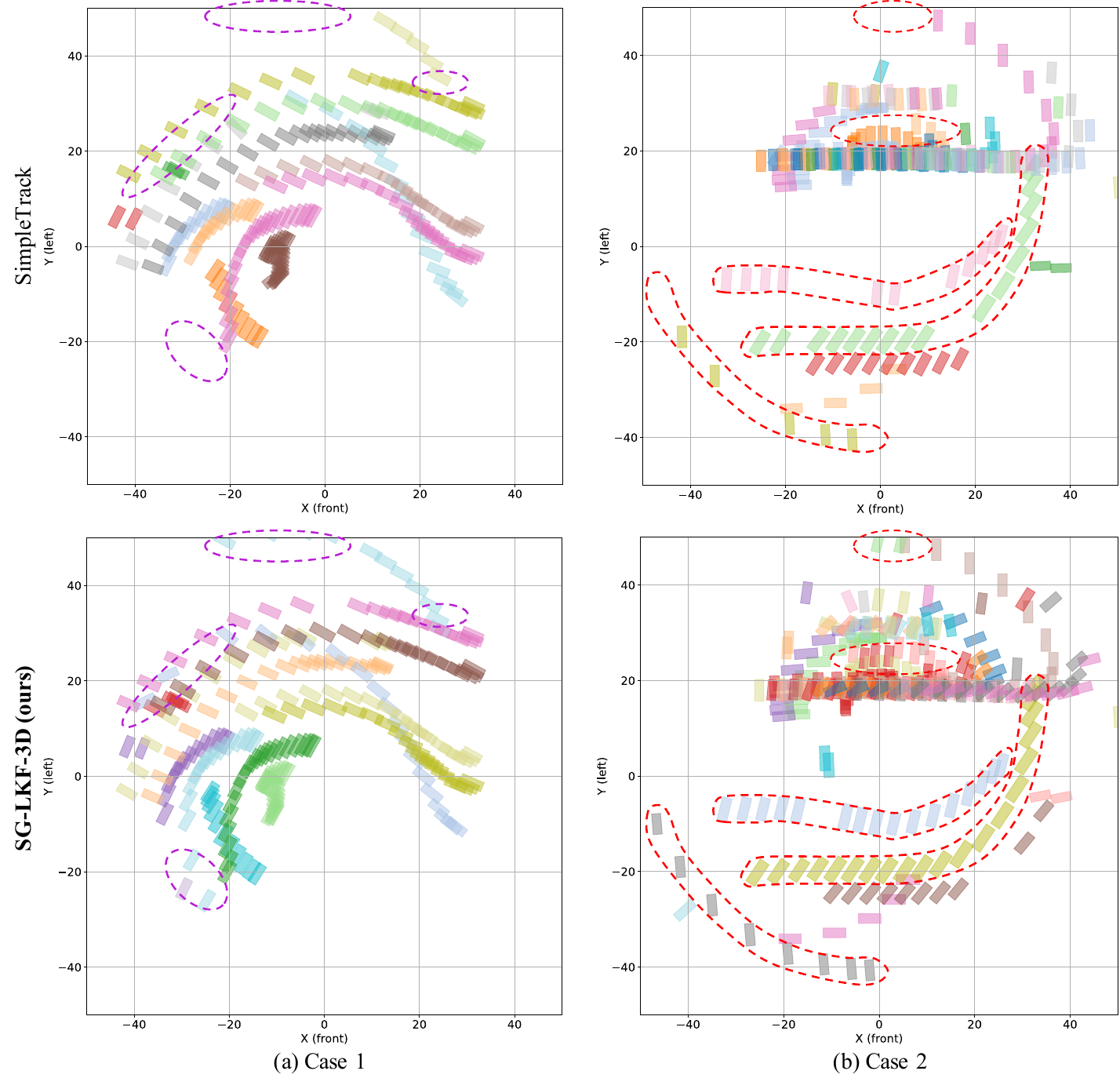} 
\caption{Qualitative comparison of SimpleTrack and SG-LKF-3D (ours) at 15$\sim$30 km/h on the nuScenes 3D MOT test set. As highlighted by the dashed boxes, SimpleTrack exhibits severe tracking failures in both case 1 and case 2. Different colors represent trajectories with unique IDs.}
\label{fig:vis_result_nus}
\end{figure}

\subsubsection{Qualitative Analysis.}
To assess SG-LKF-2D’s tracking stability under static and high-speed conditions, we qualitatively compare it with the baseline at 0 km/h and 60 km/h (see Fig.~\ref{fig:vis_result_kitti}). At 0 km/h, the BoT-SORT exhibits occlusion-induced misses and fragmented trajectories, while at 60 km/h, it suffers from association errors and poor localization. In contrast, SG-LKF-2D maintains stable and accurate tracking across different speeds.
Additionally, we further evaluate 20-frame vehicle trajectories in BEV on the nuScenes dataset (see Fig.~\ref{fig:vis_result_nus}). 
In both case 1 and case 2, SG-LKF-3D consistently demonstrates improved trajectory stability and continuity, with markedly fewer interruptions, trajectory losses, and ID switches compared to SimpleTrack.
Additional examples are provided in \textit{Appendices C and D.}

\subsection{Discussions and Limitations.}
Although SG-LKF leverages ego-vehicle speed to enhance multi-object tracking, its evaluation is limited to moderate speed variations due to existing public dataset constraints, leaving its effectiveness under high-speed (100 km/h) or irregular motion scenarios unverified.
Furthermore, the lightweight MSNet, though efficient, limits performance under frequent occlusions and dense interactions. Future work will explore advanced architectures to further improve robustness and generalization.

%

\section{Conclusion}

In this paper, we are the first to introduce ego-vehicle speed as a key prior for ego-motion modeling, effectively compensating for reference frame shifts caused by ego-vehicle dynamics. 
Building on this insight, we propose a SG-LKF, which employs a speed-aware MSNet to predict key Kalman Filter parameters, effectively correcting the bias of static motion models. Additionally, trajectory, semantic, and positional consistency losses are introduced to improve tracking stability and continuity.
Experimental results demonstrate that SG-LKF achieves the best performance and ranks first among vision-based methods on the KITTI 2D MOT benchmark. Benefiting from its high compatibility with the Kalman filter framework, SG-LKF can be seamlessly integrated into other Kalman-based tracking systems, achieving leading results on KITTI and nuScenes 3D MOT.


\bibliography{aaai2026}

\end{document}